\documentclass[lettersize,journal]{IEEEtran}
\usepackage{amsmath,amsfonts}
\usepackage{algorithmic}
\usepackage{algorithm}
\usepackage{array}
\usepackage[caption=false,font=normalsize,labelfont=sf,textfont=sf]{subfig}
\usepackage{textcomp}
\usepackage{stfloats}
\usepackage{url}
\usepackage{verbatim}
\usepackage{graphicx}
\usepackage{cite}

\usepackage{xcolor}  
\usepackage{graphicx}
\usepackage{float}
\usepackage{cite}
\usepackage{tabu}                     
\usepackage{multirow}                 
\usepackage{multicol}                 
\usepackage{multirow}                
\usepackage{float}                    
\usepackage{makecell}                 
\usepackage{booktabs}                 

\usepackage{soul, color, xcolor}

\begin{document}

\title{FedDiff: Diffusion Model Driven Federated Learning for Multi-Modal and Multi-Clients }

\author{Daixun Li, Weiying Xie, \IEEEmembership{Senior Member,~IEEE}, Zixuan Wang, Yibing Lu, Yunsong Li, \IEEEmembership{Member,~IEEE}, and Leyuan Fang, \IEEEmembership{Senior Member,~IEEE}
\thanks{This work was supported in part by the National Natural Science Foundation of China under Grant 62121001, U22B2014, and the Youth Talent Promotion Project of China Association for Science and Technology under Grant 2020QNRC001, in part by the Fundamental Research Funds for the Central Universities under Grant QTZX23048.}      
\thanks{D. Li, W. Xie, Z. Wang, Y. Lu, Y. Li are with the State Key Laboratory of Integrated Services Networks, Xidian University, Xi'an 710071, China (e-mail: ldx@stu.xidian.edu.cn; wyxie@xidian.edu.cn; ysli@mail.xidian.edu.cn;).}
\thanks{L. Fang is with the College of Electrical and Information Engineering, Hunan University, Changsha 410082, China (e-mail: fangleyuan@gmail.com).} }

\markboth{IEEE Transactions on Circuits and Systems for Video Technology}%
{Shell \MakeLowercase{\textit{et al.}}: A Sample Article Using IEEEtran.cls for IEEE Journals}

\IEEEpubid{0000--0000/00\$00.00~\copyright~2023 IEEE}

\maketitle

\begin{abstract}
With the rapid development of imaging sensor technology in the field of remote sensing, multi-modal remote sensing data fusion has emerged as a crucial research direction for land cover classification tasks. While diffusion models have made great progress in generative models and image classification tasks, existing models primarily focus on single-modality and single-client control, that is, the diffusion process is driven by a single modal in a single computing node. To facilitate the secure fusion of heterogeneous data from clients, it is necessary to enable distributed multi-modal control, such as merging the hyperspectral data of organization A and the LiDAR data of organization B privately on each base station client. In this study, we propose a multi-modal collaborative diffusion federated learning framework called FedDiff. Our framework establishes a dual-branch diffusion model feature extraction setup, where the two modal data are inputted into separate branches of the encoder. Our key insight is that diffusion models driven by different modalities are inherently complementary in terms of potential denoising steps on which bilateral connections can be built. Considering the challenge of private and efficient communication between multiple clients, we embed the diffusion model into the federated learning communication structure, and introduce a lightweight communication module. Qualitative and quantitative experiments validate the superiority of our framework in terms of image quality and conditional consistency. To the best of our knowledge, this is the first instance of deploying a diffusion model into a federated learning framework, achieving optimal both privacy protection and performance for heterogeneous data. Our FedDiff surpasses existing methods in terms of performance on three multi-modal datasets, achieving a classification average accuracy of 96.77$\%$ while reducing the communication cost.
\end{abstract}


\section{Introduction}
\IEEEPARstart{W}{ith} the rapid development of deep learning technology in the field of remote sensing \cite{song2022multiple, wang2023decomposed}, imaging satellite data has made significant progress in ground object classification tasks, opening up new possibilities for land use and urban mapping. For instance, hyperspectral satellites enable the collection of continuous spectral features, thereby facilitating high-precision land cover classification. However, single-mode satellite data has limitations. Hyperspectral satellites often struggle to provide accurate structural information, especially in the presence of extensive cloud coverage or atmospheric interference. This is where LiDAR data becomes invaluable, as it can offer surface height information, ground structure details, and other complementary data, enhancing the comprehensiveness and accuracy of remote sensing applications. 
Due to the ever-expanding availability and resolution of satellite remote sensing data, more and more Earth observation satellites are able to collect multi-modal images of the same area simultaneously \cite{ma2023mbsi}.
By fusing information from different data modalities, a better understanding and analysis of surface characteristics can be achieved, thereby greatly benefiting earth observation, environmental monitoring, resource management, and disaster response efforts.

\begin{figure}[t]
	\centering
	\includegraphics[scale=0.912]{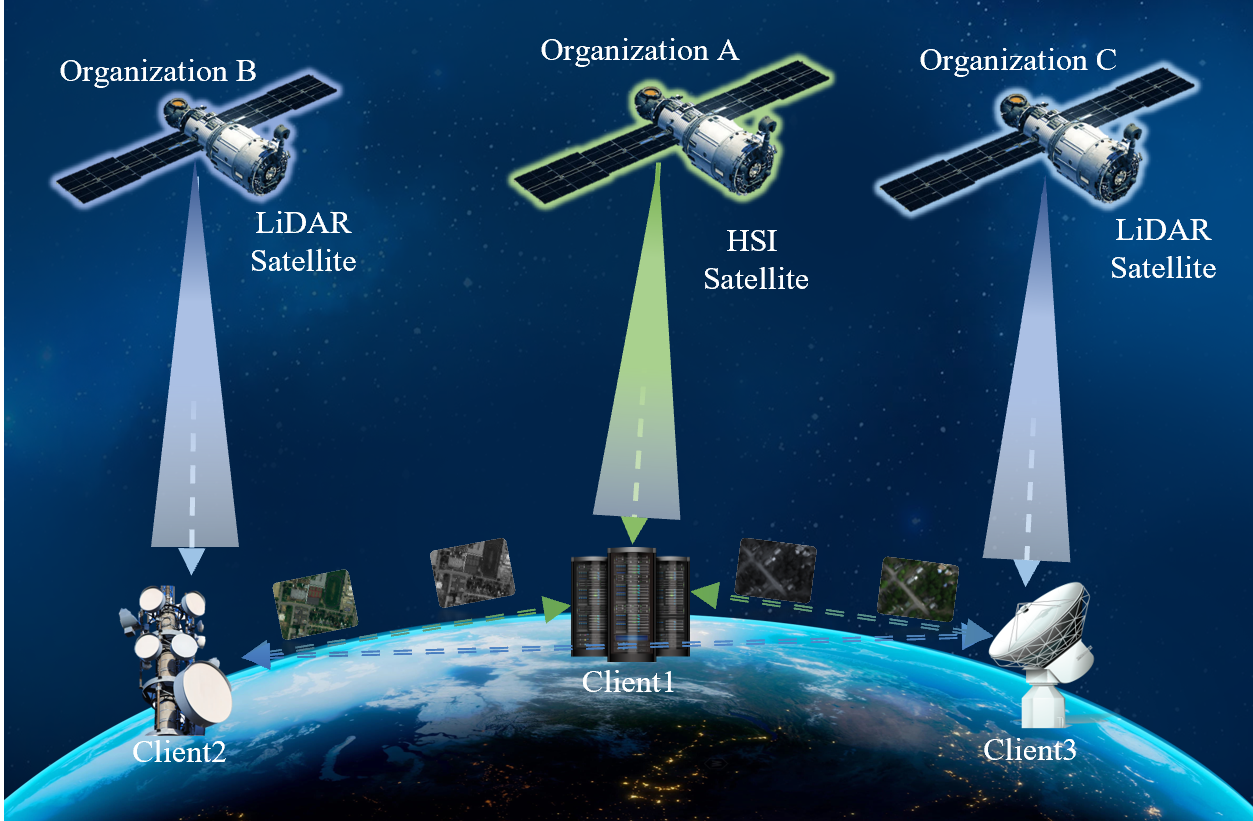}
	\centering
	\vspace{-0.1in}
	\caption{FedDiff framework at multi-clients and multi-modal network.  It includes HSI clients and LiDAR clients. There is no need to interact with original data between clients, and the global model is updated by transmitting intermediate features to achieve multi-modal data fusion.}
	\vspace{-0.1in}
	\label{intro}
\end{figure}
\IEEEpubidadjcol   
There has been noteworthy progress in the fusion of multi-modal remote sensing data for remote sensing surface object classification. A more compact feature representation can be obtained by combining the spectral information from hyperspectral satellites with elevation information from LiDAR. In the context of multi-modal data fusion methods, convolutional neural networks (CNNs) are often employed as they effectively capture the local structural information present in images. For instance, Ding \emph{et al.} \cite{DingGlobal} proposed a novel global-local conversion network for the joint classification of HSI and LiDAR data. Their approach fully mines and combines the complementary information of multi-modal data and local/global spectral-spatial information. Afterward, some advanced deep learning algorithms were used in multi-modal data fusion research.
Xiong \emph{et al.} \cite{xiong2022interpretable} proposed an interpretable fusion siamese network (IFSN) to solve the multi-modality remote sensing ship image retrieval (MRSSIR). 
Roy \emph{et al.} \cite{RoyMultimodal} introduced a multi-modal fusion transformer network that leverages multi-modal data as external classification markers for the transformer encoder to enhance generalization capabilities. Additionally, Zhao \emph{et al.} \cite{zhao2023ddfm} presented a fusion algorithm based on the denoising diffusion probability model. Their method formulates the fusion task as a conditional generation problem within the DDPM sampling framework, yielding impressive results in infrared-visible light image fusion. By combining different imaging techniques in remote sensing, various properties of the Earth's surface, such as spectral radiation, reflectance, height information, texture structure, and spatial features, can be captured. Leveraging multiple modalities enables a more detailed and precise description of the scene, which is not achievable through single-modality data alone \cite{wu2016parallel}. However, there are still two problems in the current multimodal fusion scheme. First, the current multimodal remote sensing lacks consideration from the characteristics of remote sensing data, that is, the current feature extraction only considers from the perspective of spaital and spectral, and lacks analysis from the perspective of frequency domain analysis. However, frequency domain filtering can enhance or remove specific ground features of remote sensing images by analyzing components of specific frequencies. Second, due to the limitation of data transmission and storage cost, the current fusion process mainly relies on performing fusion and computation on a single server.

As shown in Fig. \ref{intro}, with the continuous development of low-orbit satellites, more and more countries and organizations are collaborating to achieve remote sensing missions for collaborative fusion among multiple satellites. However, in multi-satellite distributed systems, certain types of data inherently contain sensitive information. Public access is limited to level-1 or higher products for spaceborne data, while level-0 data is subject to privacy protection laws \cite{li2022Simultaneous}. Several jurisdictions have implemented data protection regulations, such as the GDPR, that impose strict restrictions on sharing privacy-sensitive data between clients and platforms \cite{UtzUninformed,liang2021source,Jegorova2022survey,yang2021model}. Traditionally, raw data is transmitted between various terrestrial base stations, posing a risk of potential exposure of sensitive information to attackers during the transmission process. This not only compromises privacy protection between satellites but also raises concerns about data integrity and security. Hence, the safe and efficient processing of source data remains a significant challenge for remote sensing multi-modal fusion. Federated learning, as a distributed machine learning technology that breaks data silos, offers a promising solution. It enables participants to achieve joint modeling by exchanging machine learning intermediate results without disclosing the underlying data, thereby providing enhanced federated feature fusion for terrestrial base station systems.

A pivotal question that arises in distributed multi-modal fusion tasks is whether it is possible to construct a unified network that simultaneously ensures private communication and efficient fusion, leading to improved classification performance. To address this issue, we conduct a systematic investigation of various structures and propose a network architecture based on three fundamental properties: (1) The fused features preserve the salient features of each modality in the spatial and frequency domains. (2) Lightweight communication facilitates the transmission of data across multiple clients. (3) Each client ensures privacy of the original data during communication. Diffusion models driven by different modalities are inherently complementary regarding the latent denoising steps, where bilateral connections can be established upon \cite{Huang2023CVPR}. Inspired by this, we expect to effectively merge the features of remote sensing HSI and LiDAR images using a dual-branch co-diffusion network. It enables us to leverage information complementarity and obtain improved results for remote sensing surface object classification. In summary, we propose a federated learning framework for the diffusion model, which enables model training and lightweight data fusion on each client while ensuring data privacy. Our primary contributions are summarized as follows:

\begin{itemize}
	\item[$\bullet$] We propose a diffusion model-driven dual-branch multi-modal learning network to explicitly consider the joint information interaction between high-dimensional manifold structures in spectral, spatial, and frequency domains, which significantly improves the feature extraction performance of remote sensing data.
	
	\item[$\bullet$] We propose a multi-modal federated learning (MFL) framework, which specifically includes a multi-modal federated interaction module and a lightweight feature decomposition module. It removes the redundant information of single modal features, promotes seamless information exchange between clients, and improves the overall efficiency of the framework.
	
	\item[$\bullet$] We construct a multi-clients base station system, simulating eight base station clients using eight computing boards, and the online data of two modalities are loaded into each client in a non-independent and identical distributed way. Through extensive experimentation, we verify that our framework outperforms existing multi-modal classification methods in terms of accuracy while maintaining lower communication cost.
\end{itemize}

\section{Related Works}

\subsection{RS Images Multi-Modality}
In the face of the ever-growing complexity and diversity in remote sensing scene representation requirements, multi-modal deep learning serves as an effective approach to enhance the accuracy of overall decision-making results by leveraging the characteristics of different modal data. The multi-modal fusion method is the core content of multi-modal deep learning technology. According to the fusion stage, it can be divided into early fusion, feature-level fusion, and late fusion methods. In order to alleviate the inconsistency problem between the original data in each modality, the representation of features can be extracted from each modality separately, and then fused at the feature level, known as feature fusion. Since deep learning essentially involves learning specific representations of features from raw data, which sometimes requires data fusion before features are extracted, both feature-level and data-level fusions are called early fusion. For example, Audebert \emph{et al.} \cite{Audebert2018} proposed an early fusion network based on the FuseNet principle to jointly learn more substantial multi-modality features. He \emph{et al.} \cite{HeMultimodal} presented a hypergraph parser to imitate guiding perception to learn intra-modal object-wise relationships. Hong \emph{et al.} \cite{HongDeep} proposed the classification of hyperspectral and LiDAR data through deep encoder-decoder network architecture. The late fusion method, also known as decision-level fusion, involves training separate deep learning models for different modalities and subsequently combining the output results from multiple models. For example, P. Bellmann \emph{et al.} \cite{BellmannUsing} proposed a sample-specific late-fusion classification architecture to improve performance accuracy. The hybrid fusion method combines the early and late fusion approaches, leveraging the strengths of both methods while also introducing increased structural complexity and training challenges to the model. In multi-modal remote sensing data fusion, network layer fusion shows better competitiveness.

Furthermore, from the perspective of model structure, the framework for remote sensing multi-modal classification can be broadly categorized into traditional models, classic Convolutional Neural Networks (CNNs), Transformer models, and diffusion models. In terms of conventional model, Guo \emph{et al.} \cite{GuoOptimal} proposed the potential of using support vector machine (SVM) and random fores to estimate forest above-ground biomass through multi-source remote sensing data.
Despite their great success, traditional methods usually exhibit poor generality under different conditions due to their limited ability to construct strong feature representations and derive scene-dependent fusion strategies \cite{luo2023multi}.
For classical convolutional networks, Fang \emph{et al.} \cite{fang2021s2enet} proposed the integration of a spatial-spectral enhancement module (S2EM) into deep neural networks to facilitate cross-modal information interaction. The S2EM comprises two components: the SpAtial Enhancement Module (SAEM) and the SpEctral Enhancement Module (SEEM). In terms of Transformer architecture, Hoffmann \emph{et al.} \cite{hoffmann2023transformer} proposed a new Synchronized Class Token Fusion (SCT Fusion) architecture, which handles different input modalities by leveraging a modality-specific attention-based Transformer encoder, and the inter-modal information exchange is carried out by synchronizing the special class token after each Transformer encoder block. Remote sensing image classification necessitates the consideration of spatial information within the image, as the spatial distribution and contextual relationships of ground objects significantly influence the classification outcomes. CNNs are widely adopted in image classification tasks due to their excellent feature extraction capabilities. However, traditional CNN models do not explicitly account for the spatial relationships between pixels, which may limit their effectiveness in remote sensing image classification. On the other hand, the transformer model primarily excels in processing sequence data, such as sentences in natural language processing. Although the Transformer model can capture long-range dependencies within sequential data, for remote sensing image classification, spatial dependencies hold greater importance. Consequently, the Transformer model is not typically considered the primary choice for this specific task.

Recently, diffusion models have garnered significant attention within the deep learning community. These models excel at generating high-quality images by simulating the intricate diffusion process that effectively restores noise-corrupted images to their pristine state. Diffusion models have been applied to various fields such as image superresolution \cite{EsserTaming}, semantic segmentation \cite{baranchuk2021label}, and classification \cite{han2022card}. In the domain of multi-modal fusion, Huang \emph{et al.} \cite{Huang2023CVPR} introduced collaborative diffusion as a novel approach for achieving multi-modal face generation and editing. Unlike traditional methods that require re-training of models, the proposed approach leverages pre-trained single-modal diffusion models to achieve multi-modal face generation and editing. Furthermore, Dif-Fusion \cite{yue2023dif} employed diffusion models to generate the distribution of multi-channel input data, thereby enhancing the capability of aggregating multi-source information and maintaining color fidelity. Despite the satisfactory outcomes achieved by these methods, it is worth noting that current diffusion models employed in remote sensing classification fail to consider the manifold structure of high-dimensional data, which can have ramifications on decision-making accuracy.

\subsection{Federated Learning}
Federated learning, a distributed machine learning framework, has gained considerable attention due to its ability to ensure privacy protection and security encryption. This framework is specifically designed for training deep learning models on remote devices or isolated data centers while preserving data localization. In the applications scenarios of federated learning, it is common for each edge device to acquire data that is highly non-independent and identically distributed (Non-IID). For instance, satellites of various types, such as hyperspectral, LiDAR, and SAR satellites, often carry sensors that capture diverse data types. Moreover, the number of datapoints across devices may vary significantly, and there may be an underlying statistical structure present that captures the relationship among devices and their associated distributions \cite{smith2017federated}. This data generation paradigm violates the independent and identically distributed (IID) assumption often used in distributed optimization. This departure from the IID assumption can potentially introduce complexities in problem modeling, theoretical analysis, and empirical evaluation of solutions \cite{li2020federated}. Several studies have focused on developing efficient Federated Learning (FL) algorithms to address non-independent and identically distributed (Non-IID) data scenarios. Notable works include FedProx \cite{FedProx}, SCAFFOLD \cite{Scaffold}, and FedNova \cite{FedNova}. FedProx \cite{FedProx} can be seen as a generalization and reparametrization of FedAvg \cite{FedAvg}. It introduced proximal terms to ensure the aggregation of partial information that has not completed calculations and tackles the challenge of data heterogeneity both theoretically and empirically. FedNova \cite{FedNova} presented a comprehensive framework for analyzing the convergence of federated heterogeneous optimization algorithms. It automatically adjusts aggregate weights and effective local steps based on local progress. However, these studies may lack generalizability in simulating Non-IID datasets and may not fully cover all variations of Non-IID scenarios \cite{Federatedlearningonnon-iiddatasilos}.

In federated networks, communication serves as a critical bottleneck \cite{bonawitz2019towards}. Due to limited bandwidth, energy, and power resources, network communication can be orders of magnitude slower compared to local computing \cite{van2009multi}. To address this, two key aspects must be considered: (1) Reducing the total number of communication rounds. (2) Minimizing the size of transmitted information per round. Compression schemes, such as enforcing sparsity and low-rank properties in the update model \cite{konevcny2016federated}, or utilizing structured random rotations for quantization \cite{konevcny2016federated}, can significantly reduce the volume of messages transmitted per round. Additionally, employing local update methods can effectively decrease the total number of communication rounds. Dai \emph{et al.} \cite{dai2015high} proposed high-performance distributed machine learning at a large scale by parameter server consistency models. Also, not every device needs to be involved in every iteration of the training process. Ren \emph{et al.} \cite{ren2020scheduling} proposed a local-global federation averaging algorithm that combines local representation learning and global model federation training to improve the model's flexibility in handling heterogeneous data. However, these methods spend much effort extracting heterogeneous data features, and most of the architectures are based on edge-side devices. Therefore, it is significant to mine the heterogeneous feature information of remote sensing images to improve classification accuracy.

\section{Methodology}

\subsection{Problem Definition}
The objective of land cover classification is to precisely categorize and identify different surface features (such as buildings, water bodies, vegetation, roads, etc.) by analyzing the distinctive characteristics and attributes present in remote sensing images. Given two images of different modalities stored on heterogeneous devices $I_1, I_2 \in \mathbb{R}^{h \times w \times c}$, where $h$ and $w$ denote the height and width of the images and $c$ represents the number of image channels. The image size is $p$ pixels, $I_1(p)$ and $I_2(p)$ represent the $p$-th pixel pair of the first and second modality. The two modalities capture the same scene with the same label information, denoted as $L \in \mathbb{R}^{h \times w \times c}$ with $N$ classification categories. The specific objectives of federated multi-modal remote sensing land classification can be defined as follows: firstly, training a new model $M(\cdot)$ on diverse devices; and secondly, performing feature fusion at specific locations. We denote the fused model as $M(I_1, I_2)$, which can effectively map input images of different modalities to a novel representation $P_{max}(I_1, I_2)$. This representation signifies the probability associated with each pixel being classified into distinct categories. By setting a maximum probability threshold $\tau$ for each category, we obtain a binary prediction map through hard classification. In this map, the values 1 and 0 represent the existence of this category and other categories respectively. This classification scheme is defined as follows:
\begin{equation}
	M\left(I_1, I_2\right)=\left\{\begin{array}{lc}
		0, & \text { if } P_{\max }\left(I_1, I_2\right)<\tau, \\
		1, & \text { otherwise. }
	\end{array}\right.
\end{equation}

\begin{figure*}[htbp]
	\centering
	\includegraphics[scale=0.6]{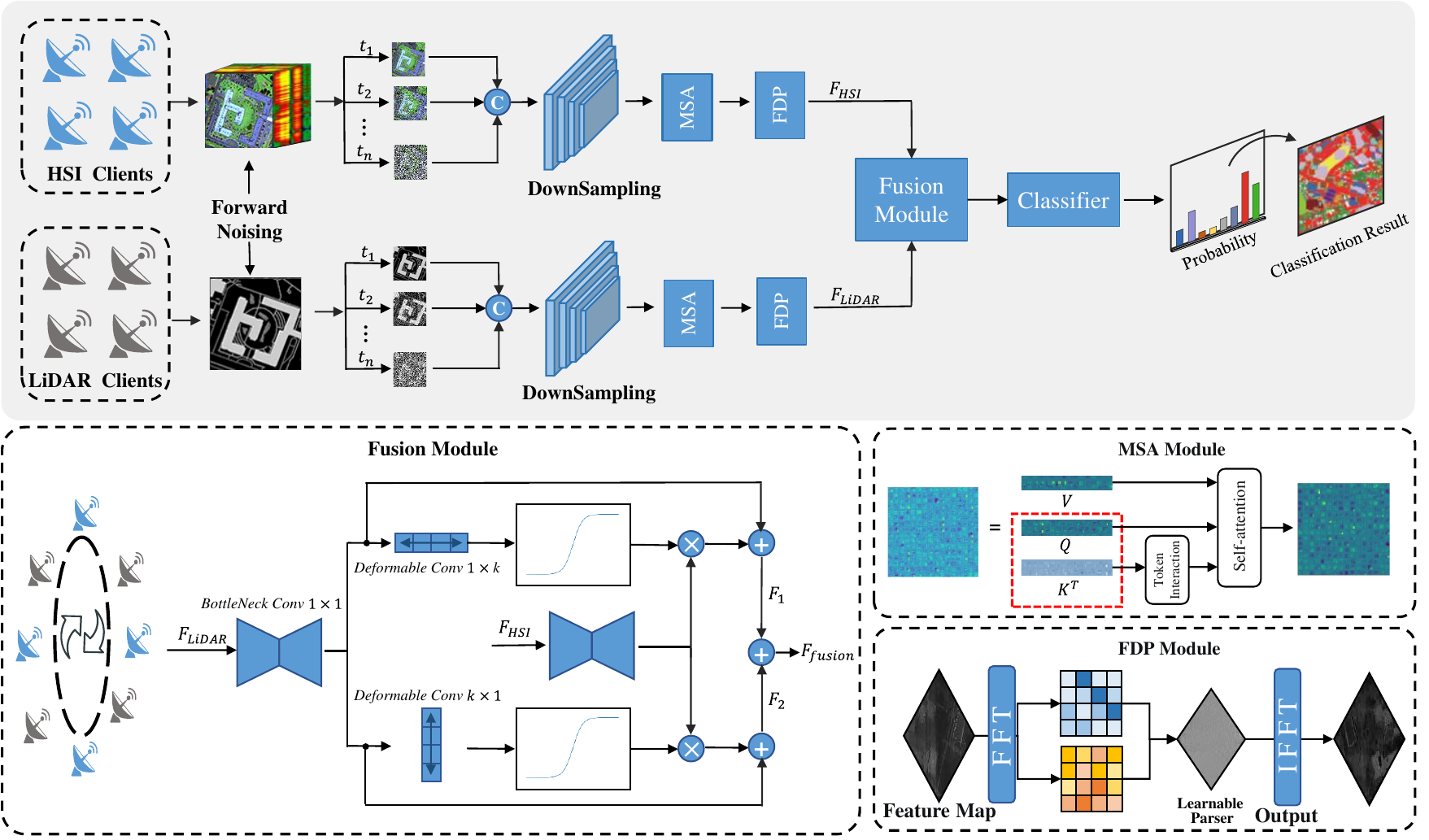}
	\centering
	\vspace{-0.1in}
	\caption{Overview of the proposed FedDiff framework.   The framework includes (1) Local Diffusion Module, (2) Global Data Fusion, and a classifier.   The architecture will be optimized with respect to the mean squared error (MSE) loss of the classifier and task-specific loss for scene classification.   In the global data fusion phase, the interactive information is propagated to each client through federated learning, in which the multi-modal federated learning communication architecture is used to reduce communication cost.}
	\vspace{-0.1in}
	\label{mainframework}
\end{figure*}

Based on this, we propose FedDiff, a diffusion model driven by federated learning. Illustrated in Fig. \ref{mainframework}, the network comprises two diffusion-driven branches, each residing on separate devices, thereby enabling federated multi-modal learning. Consequently, the fused feature map can be defined as:
\begin{equation}
	P_{\max }\left(I_1, I_2\right)=\Psi\left(I_1, I_2 \mid \theta_1, \theta_2\right),
\end{equation}
where $\Psi(\cdot)$ represents a nonlinear target model, which converts the image space into a classification space, and $\theta_1$ and $\theta_2$ respectively represent the parameters corresponding to the two branches.

\subsection{Multi-Modal Diffusion Theory}

\subsubsection{Diffusion Theory}
In order to improve the learning of the joint latent structure in high-dimensional remote sensing images, we add noise to the image in the forward diffusion process. This noise is then predicted using an improved U-Net network, allowing for the removal of noise added during the forward process in the subsequent reverse process.

Taking inspiration from the principles of non-equilibrium thermodynamics, we consider the forward diffusion process as a parameterized Markov chain \cite{sohl2015deep, ho2020denoising}. In essence, this process involves iteratively overlaying Gaussian noise onto the original image, denoted as $x_0$, until generating the final random noise $x_t$. Thus, the addition of noise in a single step can be expressed as follows:
\begin{equation}
	\mathbf{x}_t=\sqrt{\alpha_t} \mathbf{x}_{t-1}+\sqrt{1-\alpha_t} \epsilon_{t-1},
	\label{eq1}
\end{equation}
where $\alpha_t$ is a very small hyperparameter, and $\epsilon_{t-1} \sim N(0,1)$ is Gaussian noise. Deriving step by step from Eq. (\ref{eq1}), we can finally get the expression of $x_t$ as follows:
\begin{equation}
	\mathbf{x}_t=\sqrt{\bar{\alpha}_t} \mathbf{x}_0+\sqrt{1-\bar{\alpha}_t} \epsilon,
	\label{eq2}
\end{equation}
where $\bar{\alpha}_t=\prod_{i=1}^t \alpha_i$, $\epsilon \sim N(0,1)$ is Gaussian noise. According to Eq. (\ref{eq2}), random noise $x_t$ can be directly generated from the input $x_0$. The final generated random noise depends on the time step $t$, the hyperparameters $\alpha_1,…, \alpha_t$, and the random noise $\epsilon$ sampled from the standard normal distribution.

In the reverse diffusion process, we employ a denoising network to progressively restore the original high-dimensional remote sensing image. At each step $t$ of the back-diffusion process, the noisy remote sensing image $x_t$ is denoised, resulting in the image $x_{t-1}$ from the preceding step. For a high-dimensional image with a given step size $t$, the conditional probability of an image with step size $t-1$ can be expressed as \cite{ho2020denoising, gedara2022remote}:
\begin{equation}
	q\left(\mathbf{x}_{t-1} \mid \mathbf{x}_t\right)=\mathcal{N}\left(\mathbf{x}_{t-1} ; \mu_\theta\left(\mathbf{x}_t, t\right), \sigma_t^2 \boldsymbol{I}\right),
\end{equation}
where $\mu_\theta\left(\mathbf{x}_t, t\right)$ and $\sigma_t^2$ denote the mean and variance of the normal distribution, respectively. $\boldsymbol{I}$ represents the normal distribution. The mean needs to be calculated by prediction noise, while the variance $\sigma_t^2$ is determined by the hyperparameters $\alpha_1,…, \alpha_t$, and its expression is:
\begin{equation}
	\begin{gathered}
		\mu_\theta\left(\mathbf{x}_t, t\right)=\frac{1}{\sqrt{\alpha_t}}\left(\mathbf{x}_t-\frac{1-\alpha_t}{\sqrt{1-\bar{\alpha}_t}} \epsilon_\theta\left(\mathbf{x}_t, t\right)\right), \\
		\sigma_t^2=\frac{1-\bar{\alpha}_{t-1}}{1-\bar{\alpha}_t}\left(1-\alpha_t\right),
	\end{gathered}
\end{equation}
where $\epsilon_\theta\left(\mathbf{x}_t, t\right)$ represents the spectral-spatial denoising network.

\subsubsection{Double Branch Spectrum-Spatial Feature Extraction}
Taking inspiration from multi-modal cross-fusion techniques, FedDiff adopts the HSI/LiDAR diffusion dual-branch framework. This approach aims to enhance the model's performance by reusing output feature maps from different modalities. Such integration proves advantageous in processing heterogeneous data obtained from various sensors within a federated environment, leading to improved intelligent interpretation ability through better information fusion.

Based on the aforementioned analysis, it is evident that training a neural network to predict the noise introduced during the forward diffusion process is crucial for noise removal in the reverse diffusion process. Each noise step size corresponds to varying levels of texture features. Consequently, we incorporate multi-scale fusion noise as the input for our model, as follows:
\begin{equation}
	\widetilde{\mathbf{x}}=\text { Concat }\left[\mathbf{x}_0, \mathbf{x}_{50}, \mathbf{x}_{100}, \mathbf{x}_{300}, \mathbf{x}_{500}, \mathbf{x}_{700}, \mathbf{x}_{900}\right]\text {.}
\end{equation}
As FedDiff adopts a dual-branch structure, we obtain fusion noise, denoted as $\widetilde{\mathbf{x}}_1$ and $\widetilde{\mathbf{x}}_2$, generated by the two sets of data as inputs to their respective branches. The noise prediction model employed in this study is built upon the U-Net \cite{saharia2022image} network structure, as illustrated in Fig. \ref{mainframework}. Initially, the multi-modal fusion noise input by the two branches undergoes down-sampling to increase the receptive field.
\begin{equation}
	\widetilde{\mathbf{x}}_{d s}=\text { DownSampling }(\widetilde{\mathbf{x}}) \text {.}
\end{equation}

Simultaneously, to capture the spatial and spectral residual information within the digital domain of the feature map, we have incorporated a spatial-spectral attention mechanism into the foundational framework of U-Net. The process can be expressed as:
\begin{equation}
	\widetilde{\mathbf{x}}_{MSA}=\text {MSA }(\widetilde{\mathbf{x}}_{d s}) \text {.}
\end{equation}
The specific procedure is as follows: Firstly, we construct the transformation matrix $(W^Q, W^K, W^V)$ to decompose the input intermediate feature $\widetilde{\mathbf{x}}_{d s}$ into three distinct tensors: $Q$, $K$, and $V$ \cite{fan2021multiscale}.
\begin{equation}
	Q_i=\widetilde{\mathbf{x}}_{ds} W_i^Q,  K_i=\widetilde{\mathbf{x}}_{ds} W_i^K,  V_i=\widetilde{\mathbf{x}}_{ds} W_i^V.
\end{equation}
Then, we adopted the most commonly used Scaled Dot-Product Attention:
\begin{equation}
	\operatorname{Attention}(Q, K, V)=\operatorname{softmax}\left(\frac{Q K^T}{\sqrt{d_k}}\right) V.
\end{equation}
Among these components, the scaling factor $1 / \sqrt{d_k}$ is employed to mitigate the issue of vanishing gradients by scaling the result of the dot product operation between the query and key tensors. The multi-head self-attention (MSA) module divides the query, key, and value tensors into $h$ parts, facilitating each part's self-attention calculations in parallel. The outputs from each head are then concatenated and linearly projected to form the final output, which can be represented as:
\begin{equation}
		Head_i=\operatorname{Attention}\left(Q_i, K_i, V_i\right), \\
\end{equation}
\begin{equation}
		MultiHead(Q, K, V)=\operatorname{Concat}\left(Head_1, \ldots, Head_h\right) W^O. \\
\end{equation}
The multi-head attention mechanism enables the model to jointly pay attention to information from different representation subspaces at different locations, effectively capturing and leveraging the inherent interdependencies among elements within the input tensor. This approach promotes enhanced feature extraction and improves the interpretation performance of high-dimensional data within the numerical domain.

Since the resolution of remote sensing images is inherently limited by the imaging distance, blurring occurs in such images. To address this challenge, we propose the utilization of the frequency domain to enhance detailed information in different modal, including texture and color information. This enhancement improves the discrimination ability of the target, namely:
\begin{equation}
	\widetilde{\mathbf{x}}_{FDP}=FDP\left(\widetilde{\mathbf{x}}_{MSA}\right).
\end{equation}
Our primary approach involves learning a parametrized filter through applying it to the Fourier-space features.
Initially, we perform a two-dimensional Fast Fourier Transform (2D-FFT) on the input features:
\begin{equation}
	M=FFT\left(\widetilde{\mathbf{x}}_{MSA}\right).
\end{equation}
We subsequently modulate the convolution operations to acquire attention maps that highlight the significance of various frequency components. In essence, convolution weights can be considered as learnable versions of frequency filters widely used in digital image processing.
\begin{equation}
	M^{\prime}=W \otimes M,
\end{equation}
where $W$ represents the trainable weight in the frequency domain. Its purpose is to learn the convolution output in the spectral space of the preceding layer of the network while simultaneously adjusting high-frequency noise and considering its spectral characteristics. The $\otimes$ symbol denotes the element-wise product operation. Finally, we return to the spatial domain by performing an Inverse Fast Fourier Transform (IFFT), yielding the final output:
\begin{equation}
	\widetilde{\mathbf{x}}_{FDP}=IFFT\left(M^{\prime}\right).
\end{equation}

Unlike traditional methods that primarily focus on spectral or spatial domains, our approach enables global adjustments of specific frequency components, thereby enhancing the discriminative information. Thus it can be learned to constrain the high-frequency component for the adaptive integration.


\subsection{Federated Multi-Modal Fusion}
In this section, we will explore the efficient sharing of heterogeneous data from various clients, minimizing communication volume during model updates, while ensuring optimal model performance. By performing feature-level fusion of remote sensing multi-modal images, we can effectively enhance image recognition capabilities by capturing richer information expression. According to the characteristics of multi-modal network layer feature transmission in remote sensing samples, we propose a multi-modal federated learning communication architecture. 

\subsubsection{Multi-modal Federated Interaction Module}

For the sake of simplicity, we illustrate the federated learning framework using two clients as an example, namely the HSI data client and the LiDAR data client. The data in the clients is Non-IID. Our feature fusion framework enables multiple parties to collaboratively train deep learning models using data from different modalities, without transmitting original data between participants. To be specific, the server firstly distributes a copy of the model to each client. Subsequently, each client performs the diffusion process using their respective stored HSI/LiDAR data, and then transmits the trained features to other clients. 

Remote sensing images of different modalities in the same scene contain different types of features, and downstream tasks are expected to utilize multi-modal features from the upstream diffusion model to optimize efficiency. To achieve this concept, we propose a feature fusion mechanism that combines the intrinsic features of different modal images through feature reuse.   This mechanism is simple yet effective. Specifically, we propose two parallel modules, each module obtains high-level features by modulating the current layer. The process can be described as follows:
\begin{equation}
	\mathcal{F}_{fusion}=\mathcal{F}_{1}+\mathcal{F}_{2},
\label{eq18}
\end{equation}
where $\mathcal{F}_{1}$ and $\mathcal{F}_{2}$ represent the output of the two parallel branches, respectively, and $\mathcal{F}_{fusion}$  denotes the feature after fusion. $\mathcal{F}_{1}$ and $\mathcal{F}_{2}$ can be calculated as follows:
\begin{equation}
	\begin{aligned}
		& \mathcal{F}_{1}=\operatorname{Sigmoid}\left(C_d\left(\mathcal{F}_{HSI}\right)\right) C_b\left(\mathcal{F}_{LiDAR}\right)+C_b\left(\mathcal{F}_{HSI}\right), \\
		& \mathcal{F}_{2}=\operatorname{Sigmoid}\left(C_l\left(\mathcal{F}_{HSI}\right)\right) C_b\left(\mathcal{F}_{LiDAR}\right)+C_b\left(\mathcal{F}_{HSI}\right),
	\end{aligned}
\label{eq19}
\end{equation}
where, $\mathcal{F}_{HSI}$ and $\mathcal{F}_{LiDAR}$ denote the HSI and LiDAR feature maps of the final output of the two diffusion branches, respectively. $C_b(\cdot)$ is suitable for the $1 \times 1$ convolutional layer structure constraining high-frequency noise, $C_d(\cdot)$ represents the $1 \times 1$ deformable convolutional layer used to extract the deep layer, $C_l(\cdot)$ represents the $1 \times 1$ variability convolutional layer used to extract the shallow layer. Given the inherent instability of the communication environment in real situations, it is not feasible to transmit the feature map in every iteration. Instead, we adopt a pragmatic approach by selecting an appropriate communication interval based on the specific situation. Our aim is to minimize the communication cost while ensuring the model's accuracy.

After the local fusion of features, the loss function of the local model is calculated accordingly:
\begin{equation}
	\mathcal{L}=\Phi(\mathcal{F}_1, \mathcal{F}_2, \mathcal{F}_{fusion}),
\end{equation}
where $\Phi(\cdot)$ represents loss function, which will be discussed in detail in the sequel. Then, each client obtains the gradient through backward propagation, and transmits and averages the gradient:
\begin{equation}
	\bar{g}=\text {AllReduce}\left(g_1, g_2\right),
\end{equation}
where $g_1,g_2$ denote the gradient information of clients. Finally, we need to perform a global model updating:
\begin{equation}
	w_{t+1}=w_t-\eta \bar{g_t},
\end{equation}
At each iteration $t$, $\bar{g_t}$ is global average gradient, $w_t$ represents the weights of the model and $\eta$ is the learning-rate.

As a typical example of vertical federated learning, remote sensing multi-modal federated learning allows the aggregation of heterogeneous data from each client in the same scene, thereby completing the aggregation of multi-source data. This facilitates the amalgamation of multi-source data. Vertical federated learning classifies based on the feature dimensions of the data and extracts relevant parts of data with different features in the same scenario, effectively improving the feature dimensions of the original dataset. To train the model without transmitting original data, FedDiff leverages network layer feature-level fusion to aggregate and average the local gradients from all clients.

\subsubsection{Lightweight Feature Decomposition Module}
In our FedDiff federated framework, we have observed that while the size of the single-modal data feature map may not be substantial, the communication cost associated with it can become relatively high as the communication cycle increases. 
To address this issue, we aim to explore federated optimization algorithms that compress feature maps exchanged between clients to improve federated efficiency.

%
Taking into account the sparse nature of remote sensing data, we incorporated a low-rank Singular Value Decomposition (SVD) that significantly reduces the requirement for transmitting single-modal feature maps. This approach involves conducting random sampling at each client to identify the subspace that captures the crucial structure of the feature map.  Subsequently, the feature matrix representation is compressed within this low-rank subspace and reconstructed upon reception after transmission. By employing this method, we effectively minimize the amount of data that needs to be transmitted while preserving the essential information contained within the feature map.

Specifically, for a two-dimensional feature map $\mathrm{X} \in \mathbb{R}^{P \times Q}$, the SVD decomposition can be expressed as:
\begin{equation}
	\mathrm{X}=U \Sigma V^T,
\end{equation}
where $U \in \mathbb{R}^{P \times t}$ and $V \in \mathbb{R}^{t \times Q}$ are orthogonal matrix, $\Sigma \in \mathbb{R}^{t \times t}$ is a diagonal matrix of non-negative diagonal elements arranged in descending order, and the diagonal elements in $\Sigma$ are singular values. In remote sensing scene classification, we make full use of the redundant characteristics of remote sensing data, perform low-rank decomposition of local features, and use dynamic approximation methods to split the diagonal matrix.  Simultaneously, the local features are reduced using a diagonal matrix and transmitted to each client for reassembly. This low-rank representation ensures that ample information is retained, facilitating near-lossless restoration during the image reconstruction process. Consequently, our approach maximizes the utilization of redundant characteristics in remote sensing data to achieve high-fidelity restoration of the image.

\begin{algorithm}[htpb]
	\caption{Our FedDiff Algorithm. Assume that there are two clients, the HSIs client and the LiDAR images client. The learning rate $\eta$.}
	\label{alg:alg1}
	\begin{algorithmic}
		\STATE 
		\STATE {\textbf{Input:} {HSIs stored in client 1 $I_1$, LiDAR images stored in client 2 $I_2$, the model that has been initialized $\mathcal{M}$.}}
		\STATE {\textbf{Output:} {Classification results.}}
		
		\FOR{each epoch}
		\FOR{each iteration}
		\STATE $\mathcal{F}_{HSI} \gets \mathcal{M}(I_1)$
		\STATE $\mathcal{F}_{LiDAR} \gets \mathcal{M}(I_2)$
		\STATE $\text{AllGather}(\mathcal{F}_{HSI},\mathcal{F}_{LiDAR})$ (SVD)
		\STATE $\mathcal{F}_{fusion}=\text{Fusion}(\mathcal{F}_{HSI}, \mathcal{F}_{LiDAR})$ (Eq.(\ref{eq18}, \ref{eq19})
		\STATE $\mathcal{L}=\Phi(\mathcal{F}_{HSI}, \mathcal{F}_{LiDAR}, \mathcal{F}_{fusion})$
		\STATE $\bar{g}=\text{AllReduce}\left(g_1, g_2\right)$
		
		\STATE $w_{t+1}=w_t-\eta\bar{g_t}$
		\ENDFOR
		\ENDFOR
	\end{algorithmic}
\end{algorithm}

\subsection{Loss Function}
The output of $\boldsymbol{\mathcal{F}_{HSI}}$, $\boldsymbol{\mathcal{F}_{LiDAR}}$ and $\boldsymbol{\mathcal{F}_{fusion}}$ passing through the global data spreading module are defined as $\boldsymbol{\mathcal{O}_{HSI}}$, $\boldsymbol{\mathcal{O}_{LiDAR}}$ and $\boldsymbol{\mathcal{O}_{fusion}}$. The loss function of the entire network is defined as follows:
\begin{equation}
	\begin{gathered}
		\mathcal{L}_{\mathrm{CE}}=-\sum_{i=1}^n \boldsymbol{\mathcal{O}}_{\mathrm{t}} \log \boldsymbol{\mathcal{O}_{fusion}}, \\
		\mathcal{L}_{\mathrm{MSE}}=\frac{1}{n} \times\left(\sum\left(\boldsymbol{\mathcal{O}_{fusion}}-\boldsymbol{\mathcal{O}}_{\mathbf{1}}\right)^2+\sum\left(\boldsymbol{\mathcal{O}_{fusion }}-\boldsymbol{\mathcal{O}}_{\mathbf{2}}\right)^2\right), \\
		\mathcal{L}=\mathcal{L}_{\mathrm{CE}}+\mathcal{L}_{\mathrm{MSE}}+\|w\|_2 .
	\end{gathered}
\end{equation}
where, $\boldsymbol{\mathcal{O}_t}$ refers to the actual value of the sample, $n$ represents the number of observed values in the dataset, and $\|w\|_2$ denotes the sum of the L2 norm, which is utilized for computing the weights of all network layers. This method effectively mitigates the issue of overfitting, which can arise from an excessive number of model parameters.

\section{Experiments and Analysis}

\subsection{Data Description}
In order to verify the superiority of FedDiff in analyzing high-dimensional multi-modal data, we selected three remote sensing multi-modal land classification datasets for verification of downstream tasks, namely the Houston2013 dataset, the Trento dataset and the MUUFL dataset.   Detailed introduction is as follows:
\subsubsection{The HSI-LiDAR Houston2013 Dataset}
The HSI-LiDAR Houston2013 dataset is a valuable earth observation dataset that combines hyperspectral and LiDAR data for in-depth studies on land cover and environmental changes. This dataset was created by CosmiQ Works team and uses high-resolution WorldView-2 satellite imagery. Collected by the ITRES CASI-1500 imaging sensor in 2013, an extensive data collection campaign was conducted in the University of Houston campus area in Texas, USA. The hyperspectral component of this dataset comprises 144 bands, covering a wide wavelength range of 0.38$\mu m$ to 1.05$\mu m$, with a spectral interval of 10$nm$. On the other hand, the LiDAR dataset consists of a single-band image. All images in the dataset have dimensions of $349 \times 1905$ pixels. The distribution of training and test samples used for classification tasks in the experiment is shown in Table \ref{tab-houstondataset}.

\begin{table}[H]
	\renewcommand{\arraystretch}{1.3}
	\setlength{\tabcolsep}{2.1mm}{
		\caption{A List of the Number of Training and Testing Samples for Each Class in Houston2013 Dataset}
		\label{tab-houstondataset}
		\centering
		\begin{tabular}{ccc|ccc}
			\toprule[1.2pt]
			\textbf{Land cover} & \textbf{Train} & \textbf{Test} & \textbf{Land cover} & \textbf{Train} & \textbf{Test} \\ 
			\midrule[1.2pt] 
			Background & 662013 & 652648 & Grass-healthy & 198 & 1053 \\ 
			Grass-stressed & 190 & 1064 & Grass-synthetic & 192 & 505 \\ 
			Tree & 188 & 1056 & Soil & 186 & 1056 \\ 
			Water & 182 & 143 & Residential & 196 & 1072 \\
			Commercial & 191 & 1053 & Road & 193 & 1059 \\ 
			Highway & 191 & 1036 & Railway & 181 & 1054 \\ 
			Parking-lot1 & 192 & 1041 & Parking-lot2 & 184 & 285 \\ 
			Tennis-court & 181 & 247 & Running-track & 187 & 473 \\ 
			\bottomrule[1.2pt]
	\end{tabular}}
\end{table}

\subsubsection{The Trento Multi-Modal Dataset}
The Trento dataset was obtained from a rural area located south of the city of Trento, Italy. It was created by researchers at the University of Trento and utilized data sourced from the COSMO-SkyMed satellite. The COSMO-SkyMed satellite system is a synthetic aperture radar (SAR) satellite jointly developed and operated by the Italian Space Agency (ASI) and the Italian Ministry of Defense. HSI data are collected by an AISA Eagle sensor, which features $600 \times 166$ pixels and 63 spectral channels spanning from 0.40 to 0.98 $\mu m$. The spatial resolution of the HSI data is 1 meter. SAR images, which provide elevation information in a single band, were collected using an Optech ALTM 3100EA sensor. The dataset encompasses six distinct feature categories. Please refer to Table \ref{Trento} for the specific sample distribution breakdown.

\begin{table}[htpb]
	\renewcommand{\arraystretch}{1.3}
	\setlength{\tabcolsep}{2.7mm}{
		\caption{A List of the Number of Training and Testing Samples for Each Class in Trento Dataset}
		\label{Trento}
		\centering
		\begin{tabular}{ccc|ccc}
			\toprule[1.2pt]
			\textbf{Land cover} & \textbf{Train} & \textbf{Test} & \textbf{Land cover} & \textbf{Train} & \textbf{Test} \\ 
			\midrule[1.2pt] 
			Background & 98781 & 70205 & Apples & 129 & 3905 \\ 
			Buildings & 125 & 2778 & Ground & 105 & 374 \\ 
			Woods & 188 & 1056 & Vineyard & 184 & 10317 \\
			Roads & 122 & 3052 & ~ & ~ & ~ \\
			\bottomrule[1.2pt]
	\end{tabular}}
\end{table}

\subsubsection{The MUUFL Gulfport Scene Dataset}
The MUUFL dataset was collected in November 2010 at the University of Southern Mississippi-Gulf Park Campus situated in Long Beach, Mississippi. HSI data were acquired by the ITRES Research Ltd. Compact Airborne Spectral Imager (CASI)-1500 sensor. This sensor consists of $325 \times 220$ pixels with 72 spectral bands (64 available bands) spanning from 375 to 1050 $nm$, with a bandwidth of 10 $nm$. The HSI data exhibits a spatial resolution of $0.54 \times 1.0 m$. LiDAR images contain elevation data in 2 bands. The dataset encompasses 11 different types of ground objects. In total, there are 53,687 ground truth pixels available for analysis. For further details regarding the distribution of testing and training samples, please refer to Table \ref{MUUFL}.

\begin{table}[htpb]
	\renewcommand{\arraystretch}{1.3}
	\setlength{\tabcolsep}{1.63mm}{
		\caption{A List of the Number of Training and Testing Samples for Each Class in Muufl Dataset}
		\label{MUUFL}
		\centering
		\begin{tabular}{ccc|ccc}
			\toprule[1.2pt]
			\textbf{Land cover} & \textbf{Train} & \textbf{Test} & \textbf{Land cover} & \textbf{Train} & \textbf{Test} \\ 
			\midrule[1.2pt] 
			Background & 68817 & 20496 & Trees & 1162 & 22084  \\ 
			Grass-Pure & 214 & 4056 & Grass-Groundsurface & 344 & 6538  \\ 
			Dirt-And-Sand & 91 & 1735 & Road-Materials & 334 & 6353  \\ 
			Water & 23 & 443 & Buildings'-Shadow & 112 & 2121  \\ 
			Buildings & 312 & 5928 & Sidewalk & 69 & 1316  \\ 
			Yellow-Curb & 9 & 174 & ClothPanels & 13 & 256  \\ 
			\bottomrule[1.2pt]
	\end{tabular}}
\end{table}

\subsection{Evaluation Metrics and Parameter Setting}

\subsubsection{Evaluation Metrics}
To assess the effectiveness of FedDiff in remote sensing feature classification tasks on various datasets, we employed four evaluation indicators: Overall Accuracy (OA), Class Accuracy (CA), Average Accuracy (AA), and Kappa coefficient. OA is a widely used metric that measures the classification accuracy of the model over the entire dataset. CA represents the model's capability to classify and recognize individual categories within the dataset. AA is an improved version of OA that defines as the average of CA. These two indexes mitigate the influence of an imbalanced number of categories and provides a more precise evaluation of the model's performance across different categories. Kappa coefficient is an evaluation index that measures the consistency between model prediction results and actual labels in classification tasks. It considers the disparity between the model's classification accuracy and random predictions, thereby eliminating the impact of chance. The specific formulas for the aforementioned indicators are as follows:
\begin{equation}
	OA=\frac{N_c}{N_a},
\end{equation}
\begin{equation}
	CA_i=\frac{N_c^i}{N_a^i},
\end{equation}
\begin{equation}
	AA=\frac{1}{C} \sum_{i=1}^C CA_i,
\end{equation}

\begin{equation}
	\text {Kappa}=\frac{O A-P_e}{1-P_e},
\end{equation}
and
\begin{equation}
	P_e=\frac{N_r^1 \times N_p^1+\cdots N_r^i \times N_p^i+\cdots+N_r^C \times N_p^C}{N_a \times N_a},
\end{equation}
where $N_c$ and $N_a$ represent the number of correctly classified samples and the total number of samples, and $N_c^i$ and $N_a^i$ represent the number of correctly classified samples and the total number of samples in each category respectively. $C$ is the total number of categories. $P_e$ is defined as the hypothetical probability of chance coincidence. In addition, in order to evaluate the effectiveness of the efficient communication module, we calculate the size of feature maps that need to be transmitted between clients of the federated framework during each iteration.
\subsubsection{Parameter Setting}
The proposed FedDiff approach is implemented using PyTorch, with a total of 8 clients. Communication between clients is realized through the NCCL backend. Each client is equipped with an NVIDIA A100 Tensor Core GPU, simulating the HSI data clients and LiDAR data clients, respectively. The data is randomly partitioned into a training set, which comprises 95$\%$ of the data, and a validation set, which contains the remaining 5$\%$.

During the training phase, the Adam optimizer with second-order momentum was used, with the momentum parameter set to 0.9. The initial learning rate is set to 0.001 and updated via the MutilStepLR ($\gamma = 0.5$) strategy with an update interval of 60 epochs. For the three remote sensing datasets, we trained for 300 epochs, the batch size was set to 64, and L2 norm regularization was used to prevent model overfitting. It is worth noting that due to the larger size and increased number of categories in the MUUFL dataset compared to others, it has fewer labels available. Therefore, we set the batch size to 128 and optimized the ReduceLROnPlateau learning rate adjustment strategy.

\begin{table}[!ht]
	\centering
	\renewcommand{\arraystretch}{1.3}
	\setlength{\tabcolsep}{2.7mm}{
		\caption{Average Results of Ablation Study for FedDiff Performed on Three Datasets}
		\begin{tabular}{cccc}
			\toprule[1.2pt]
			\textbf{Method} & \textbf{OA(\%)} & \textbf{AA(\%)} & \textbf{Kappa(\%)} \\
			\midrule[1.2pt]	   
			\textbf{U-Net} & 94.68 & 91.23 & 94.47 \\ 
			\textbf{Local} & 96.73 & 93.65 & 95.99 \\ 
			\textbf{Non-MFL} & 96.77 & 93.69 & 96.08 \\
			\textbf{FedDiff}  & \textbf{97.16} & \textbf{93.90} & \textbf{96.45} \\ 
			\bottomrule[1.2pt]
		\end{tabular}
		\label{ablation}}
\end{table}

\subsection{Ablation Study}
In this section, we conduct a series of ablation experiments to assess the effectiveness of each module in FedDiff. Specifically, we verified the improved U-Net network and the efficient communication framework in FedDiff separately. In detail, the following four scenarios are designed: (1) Multi-modal fusion performed locally using the original U-Net network. (2) Multi-modal fusion performed locally using our improved U-Net network. (3) Our FedDiff without MFL framework. (4) Our FedDiff method. The results are summarized in Table \ref{ablation}.

\subsubsection{Validation of the Improved U-Net}
By comparing the experimental results in rows 2 and 3 of Table \ref{ablation}, it is evident that the multi-modal dual-branch diffusion fusion network employed in FedDiff outperforms the original U-Net, demonstrating significant improvements in the classification results across all three datasets. The average OA accuracy and AA accuracy on these datasets are 96.73$\%$ and 93.65$\%$, respectively, marking an improvement of 2.05$\%$ and 2.42$\%$ compared to the original U-Net. Additionally, the Kappa coefficient shows a 1.52$\%$ improvement, further validating the effectiveness of the U-Net network.

\begin{figure}[htbp]
	\centering
	\includegraphics[scale=0.09]{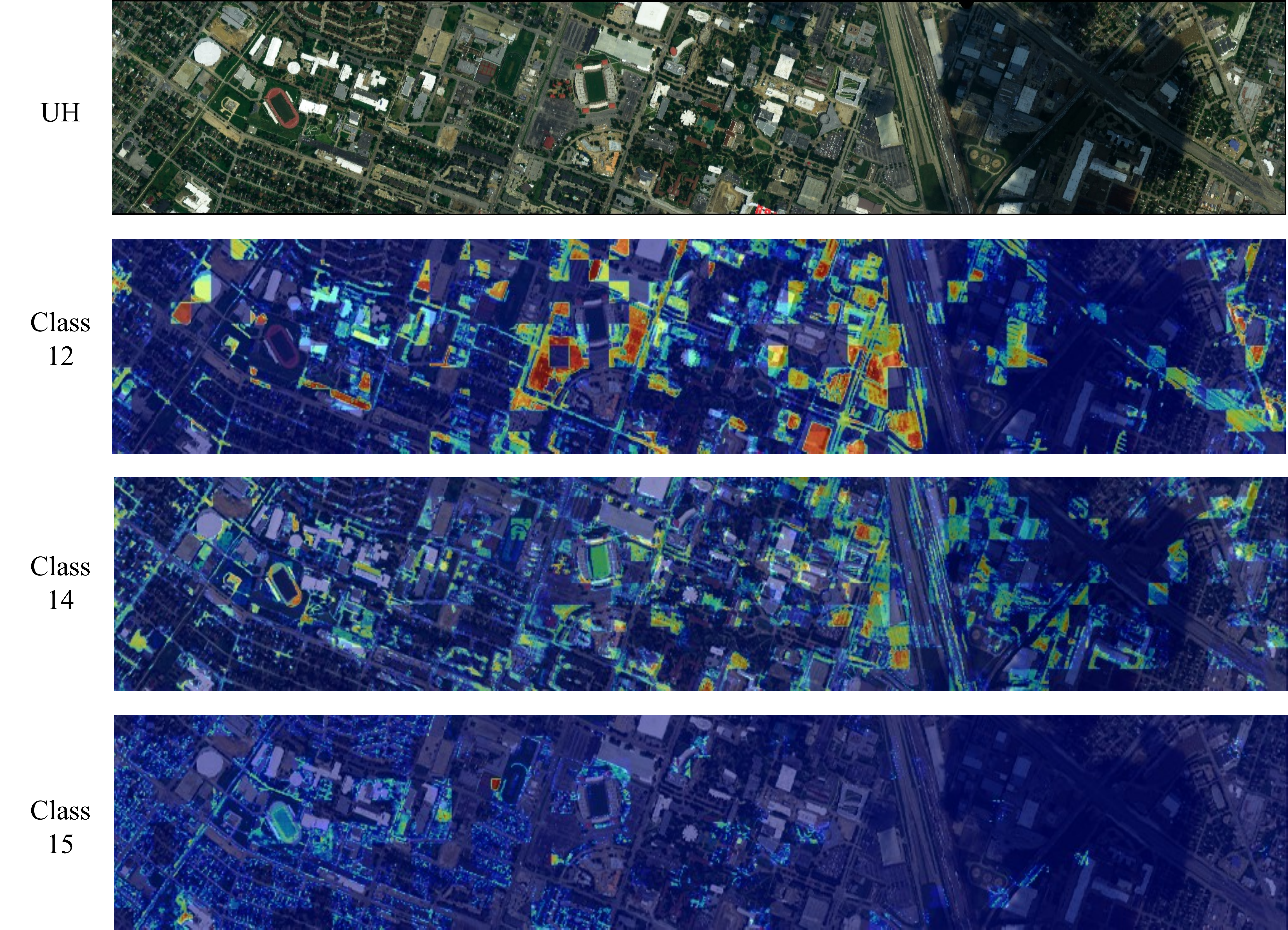}
	\centering
	\vspace{-0.1in}
	\caption{Visualization of classification results of individual categories, specifically include categories 12 (Park lot 1), 14 (Tennis court), and 15 (Running track) on the Houston2013 dataset.}
	\vspace{-0.1in}
	\label{hotmap}
\end{figure}

\subsubsection{Validation of the MFL Framework}
To validate the feasibility of multi-node multi-modal feature fusion, we deployed the aforementioned dual-branch network across 8 clients, each of which is equipped with NVIDIA A100 Tensor Core GPU and uses Pytorch platform, using NCCL as communication backend. These clients simulated eight hyperspectral imaging satellites and eight LiDAR imaging satellites and were randomly divided into a training set containing 95$\%$ data and a validation set containing 5$\%$ data. 
By comparing the data in rows 3 and 4 of Table \ref{ablation}, it is evident that distributing the network across multiple nodes does not compromise the classification accuracy. Then, to verify the effect of the efficient communication module on the accuracy of the model, ablation experiments were conducted on the MFL framework. The results in Table \ref{ablation} demonstrate that the inclusion of the MFL framework lead to an improvement in classification accuracy. This improvement can be attributed to the utilization of low-rank decomposition for the features. High-dimensional remote sensing features often contain redundancy, and by employing low-rank decomposition, redundant information can be effectively eliminated, thereby preventing over-fitting of the model.

\begin{table}[!ht]
	\centering
	\renewcommand{\arraystretch}{1.3}
	\setlength{\tabcolsep}{2.7mm}{
		\caption{Comparison of Communication Cost between FedDiff without MFL Framework and FedDiff}
		\begin{tabular}{cccc}
			\toprule[1.2pt]
			\textbf{Method} & \textbf{Bytes} & \textbf{Communication Cost} &\\
			\midrule[1.2pt]
			\textbf{Non-MFL} & 3227872 & 12.313MB & \\ 
			\textbf{FedDiff} & 1394329 & 5.318MB & 2.315$\times$\\
			\bottomrule[1.2pt]
		\end{tabular}
		\label{federated}}
\end{table}

\begin{table*}[htbp]
	\small
	\renewcommand{\arraystretch}{1.2}
	\centering
	\setlength{\tabcolsep}{4.35mm}{
		\caption{OA, AA and Kappa Coefficient on the Houston2013 Dataset (in $\%$) by Considering HSI and LiDAR Data}
		\label{tab-vsh}	
		\begin{tabular}{c|c|c|c|c|c|c|c|c|c}
			\toprule[1.2pt]
			\textbf{No.} & \textbf{Class} & \textbf{SVM} & \textbf{CNN-2D} & \textbf{RNN} & \textbf{Cross} & \textbf{CALC} & \textbf{ViT} & \textbf{MFT} & \textbf{FedDiff} \\ 
			\cmidrule(lr){1-10}
			1 & Healthy grass & 79.87 & \textbf{88.86} & 79.23 & 80.06 & 78.63 & 82.59 & 82.34 & 86.70 \\ 
			2 & Stressed grass  & 79.14 & 86.18 & 81.42 & \textbf{100.00} & 83.83 & 82.33 & 88.78 & 98.68 \\ 
			3 & Synthetis grass & 61.25 & 45.35 & 38.75 & \textbf{98.22} & 93.86 & 97.43 & 98.15 & 95.64 \\ 
			4 & Tree & 85.23 & 75.47 & 88.35 & 95.83 & 86.55 & 92.93 & 94.35 & \textbf{99.43} \\ 
			5 & Soil & 87.22 & 87.12 & 90.91 & 99.43 & 99.72 & 99.84 & 99.12 & \textbf{100.00} \\ 
			6 & Water & 61.54 & 72.03 & 77.39 & 94.41 & 97.90 & 84.15 & 99.30 & \textbf{100.00} \\ 
			7 & Residential & 81.53 & 68.13 & 65.76 & \textbf{95.15} & 91.42 & 87.84 & 88.56 & 90.95 \\ 
			8 & Commercial & 18.80 & 23.36 & 35.61 & 92.02 & 92.88 & 79.93 & 86.89 & \textbf{96.49} \\ 
			9 & Road & 68.27 & 59.05 & 72.05 & \textbf{98.68} & 87.54 & 82.94 & 87.91 & 92.82 \\ 
			10 & Highway & 48.46 & 32.37 & 34.04 & 60.71 & 68.53 & 52.93 & 64.70 & \textbf{95.08} \\ 
			11 & Railway & 38.71 & 43.33 & 36.78 & 88.71 & 93.36 & 80.99 & \textbf{98.64} & 97.72 \\ 
			12 & Park lot 1 & 57.54 & 35.57 & 58.44 & 81.94 & 95.10 & 91.07 & 94.24 & \textbf{96.45} \\ 
			13 & Park lot 2 & 65.96 & 52.28 & 52.61 & 91.23 & \textbf{92.98} & 87.84 & 90.29 & 90.18 \\ 
			14 & Tennis court & 83.40 & 70.58 & 81.24 & 97.57 & \textbf{100.00} & \textbf{100.00} & 99.73 & \textbf{100.00} \\ 
			15 & Running track & 48.41 & 45.24 & 40.94 & 98.73 & 99.37 & \textbf{99.65} & 99.58 & 99.58 \\ 
			\midrule[1.2pt]
			~ & OA(\%) & 64.18 & 59.06 & 62.61 & 90.34 & 88.97 & 85.05 & 89.80 & \textbf{95.62} \\ 
			~ & AA(\%) & 64.36 & 58.99 & 62.24 & 91.51 & 90.78 & 86.83 & 91.54 & \textbf{95.98} \\ 
			~ & Kappa(\%) & 61.30 & 55.70 & 59.64 & 89.53 & 88.06 & 83.84 & 88.93 & \textbf{92.25} \\ 
			\bottomrule[1.2pt]
	\end{tabular}}
\end{table*}

Furthermore,
as depicted in Table \ref{federated}, the inclusion of the MFL framework leads to a significant reduction in communication volume, decreasing from 12.313MB to 5.318MB. The communication cost was reduced by more than twice, effectively enhances communication efficiency among different clients. Particularly in conditions with limited communication capacity, this reduction in communication overhead proves highly advantageous. In summary, FedDiff successfully achieves a favorable trade-off between communication cost and classification accuracy.

\begin{figure*}[htbp]
	\centering
	\includegraphics[scale=0.58]{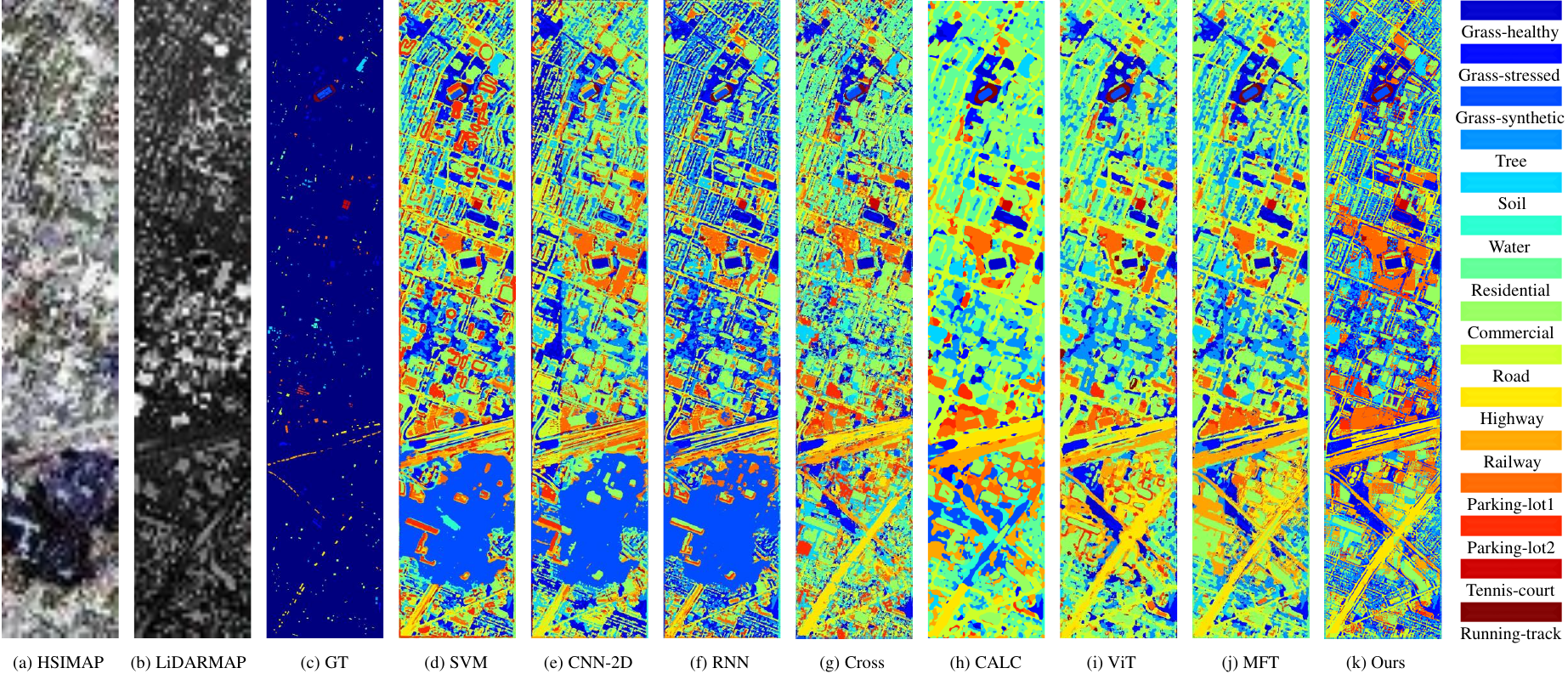}
	\centering
	\vspace{-0.2in}
	\caption{Visualization of false-color HSI and LiDAR images using different comparison methods based on the Houston2013 dataset.}
	\vspace{-0.1in}
	\label{vis-h}
\end{figure*}

\begin{table*}[htbp]
	\small
	\renewcommand{\arraystretch}{1.2}
	\centering
	\setlength{\tabcolsep}{4.5mm}{
		\caption{OA, AA and Kappa Coefficient on the Trento dataset (in $\%$) by Considering HSI and LiDAR Data}
		\label{tab-vst}	
		\begin{tabular}{c|c|c|c|c|c|c|c|c|c}
			\toprule[1.2pt]
			\textbf{No.} & \textbf{Class} & \textbf{SVM} & \textbf{CNN-2D} & \textbf{RNN} & \textbf{Cross} & \textbf{CALC} & \textbf{ViT} & \textbf{MFT} & \textbf{FedDiff} \\ 
			\cmidrule(lr){1-10}
			1 & Apple trees & 97.44 & 96.98 & 91.75 & \textbf{99.87} & 98.62 & 90.87 & 98.23 & 97.98 \\ 
			2 & Buildings & 98.12 & 97.56 & 99.47 & 98.63 & \textbf{99.96} & 99.32 & 99.34 & 99.13 \\ 
			3 & Ground & 56.15 & 55.35 & 79.23 & \textbf{98.93} & 72.99 & 92.25 & 94.92 & 97.82 \\
			4 & Woods & 97.53 & 99.66 & 99.58 & 99.22 & \textbf{100.00} & 99.63 & 99.45 & \textbf{100.00} \\ 
			5 & Vineyard & 98.13 & 99.56 & 98.39 & 98.05 & 99.44 & 98.23 & 99.72 & \textbf{100.00} \\ 
			6 & Roads & 78.96 & 76.91 & 85.86 & 89.42 & 88.76 & 84.04 & 91.52 & \textbf{97.78} \\ 
			\midrule[1.2pt]
			~ & OA(\%) & 95.33 & 96.14 & 96.43 & 97.82 & 98.11 & 94.62 & 97.76 & \textbf{99.38} \\ 
			~ & AA(\%) & 87.72 & 87.67 & 92.38 & 97.35 & 93.30 & 91.33 & 95.91 & \textbf{98.79} \\ 
			~ & Kappa(\%) & 93.76 & 94.83 & 95.21 & 97.09 & 97.46 & 92.81 & 97.00 & \textbf{99.18} \\ 
			\bottomrule[1.2pt]
	\end{tabular}}
\end{table*}

\subsection{Comparisons with Previous Methods}
The accuracy performance of our proposed model, along with other comparison method, on the Houston2013, Trento, and MUUFL datasets can be found in Tables \ref{tab-vsh}, \ref{tab-vst}, and \ref{tab-vsm}. The best results are highlighted in bold. In our comparative analysis, we considered both classic deep learning techniques such as SVM \cite{SVM}, CNN-2D \cite{CNN2D}, and RNN \cite{RNN}, and mainstream remote sensing multi-modal methods including Cross \cite{Cross}, CALC \cite{DingGlobal}, ViT \cite{ViT}, and MFT \cite{RoyMultimodal}. Based on our evaluation results, our method excels not only in achieving federated lightweight communication but also surpasses other methods in terms of OA, AA, and Kappa coefficient across other classification tasks. Specifically, on the Houston2013 dataset, FedDiff demonstrates superior performance, outperforming other mainstream methods in all three indicators. Additionally, it achieves the highest classification accuracy among the seven categories. Compared with the classic deep learning method, FedDiff fully demonstrates the advantages of its large model. Our method shows the huge advantages in OA, AA, and Kappa coefficients, where the improvements in OA, AA, and Kappa coefficients all exceed 30$\%$. When compared to the state-of-the-art MFT algorithm in the Transformer, our proposed method demonstrates significant enhancements. Specifically, OA is significantly improved by 5.82$\%$, AA is improved by 4.44$\%$, and Kappa coefficient is improved by 3.32$\%$. These results indicate that our proposed method excels even when the model size is comparable, effectively enhancing the accuracy of remote sensing multi-modal land classification. Tables \ref{tab-vst} and \ref{tab-vsm} present the classification results for the Trento and MUUFL dataset, respectively. On the Trento dataset, most of the compared methods achieve an accuracy of over 95$\%$, and the differences in performance among these methods are not significant. However, our proposed method outperforms all current mainstream methods with an OA of 99.38$\%$, an AA of 98.79$\%$, and Kappa coefficient of 99.18$\%$. As shown in Table \ref{tab-vsm}, on the MUUFL dataset, FedDiff demonstrates notable improvements compared to the previous leading method (MFT). The OA, AA, and Kappa coefficients show increases of 0.98$\%$, 4.81$\%$, and 1.31$\%$, respectively.

\begin{figure*}[htbp]
	\centering
	\includegraphics[scale=0.58]{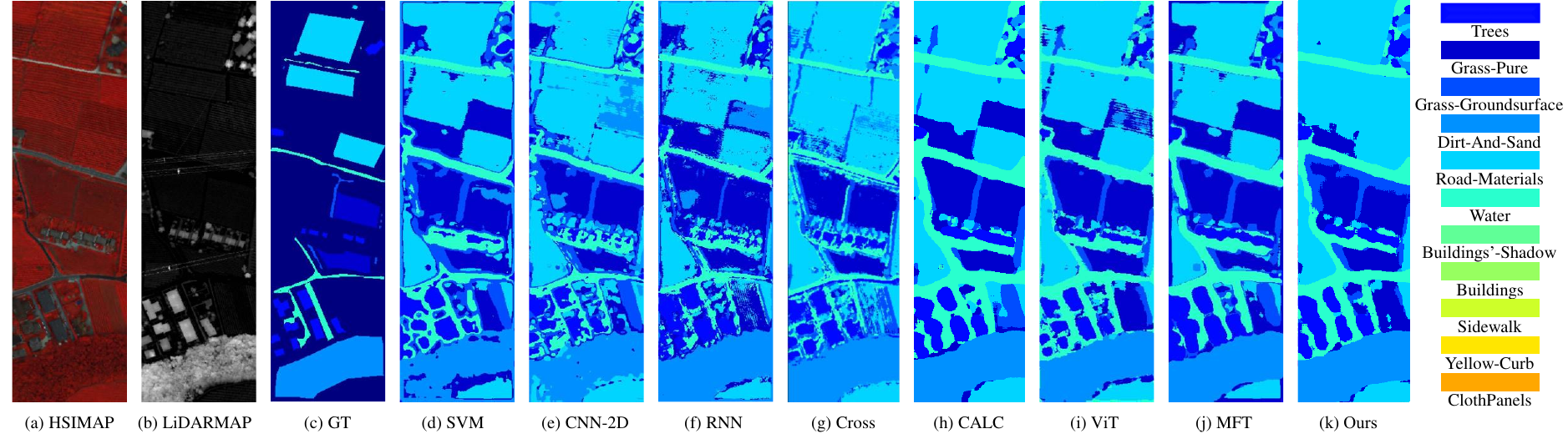}
	\centering
	\vspace{-0.2in}
	\caption{Visualization of false-color HSI and LiDAR images using different comparison methods based on the Trento dataset.}
	\vspace{-0.1in}
	\label{vis-t}
\end{figure*}

Overall, FedDiff achieves comprehensive performance improvements on three remote sensing multi-modal datasets. It effectively reduces the communication volume within the federated framework while maintaining superior classification accuracy compared to existing methods. This makes it an effective solution for feature fusion and feature extraction in federated heterogeneous high-dimensional remote sensing data scenarios.

\begin{table*}[htpb]
	\small
	\renewcommand{\arraystretch}{1.2}
	\centering
	\setlength{\tabcolsep}{4.1mm}{
		\caption{OA, AA and Kappa Coefficient on the MUUFL Dataset (in $\%$) by Considering HSI and LiDAR Data}
		\label{tab-vsm}	
		\begin{tabular}{c|c|c|c|c|c|c|c|c|c}
			\toprule[1.2pt]
			\textbf{No.} & \textbf{Class} & \textbf{SVM} & \textbf{CNN-2D} & \textbf{RNN} & \textbf{Cross} & \textbf{CALC} & \textbf{ViT} & \textbf{MFT} & \textbf{FedDiff} \\ 
			\cmidrule(lr){1-10}
			1 & Trees & 96.63 & 95.79 & 95.84 & \textbf{98.55} & 97.31 & 97.85 & 97.90 & 98.28 \\ 
			2 & Mostly grass & 59.25 & 72.76 & 81.93 & 65.85 & \textbf{93.00} & 76.06 & 92.11 & 91.42 \\
			3 & Mixed ground surface & 81.46 & 78.92 & 80.47 & 79.24 & 91.57 & 87.58 & 91.80 & \textbf{92.21} \\ 
			4 & Dirt and sand & 73.54 & 83.59 & 87.01 & 69.22 & 95.10 & 92.05 & 91.59 & \textbf{95.27} \\ 
			5 & Road & 83.79 & 78.29 & 90.65 & \textbf{97.12} & 95.91 & 94.71 & 95.60 & 95.78 \\  
			6 & Water & 15.35 & 50.34 & 54.25 & 60.95 & \textbf{99.32} & 82.02 & 88.19 & 96.84 \\  
			7 & Building shadow & 77.04 & 79.70 & 81.24 & 64.55 & \textbf{92.69} & 87.11 & 90.27 & 90.52 \\ 
			8 & Building & 86.94 & 71.95 & 88.39 & 90.30 & 98.45 & 97.60 & 97.26 & \textbf{98.55} \\  
			9 & Sidewalk & 21.28 & 43.92 & 60.54 & 32.75 & 51.60 & 57.83 & 61.35 & \textbf{73.63} \\  
			10 & Yellow curb & 0.00 & 12.45 & 26.44 & 0.00 & 0.00 & \textbf{31.99} & 17.43 & 26.44 \\ 
			11 & Cloth panels & 62.89 & 26.82 & 87.50 & 43.36 & 0.00 & 58.72 & 72.79 & \textbf{90.23} \\  
			\midrule[1.2pt]
			~ & OA(\%) & 84.24 & 83.40 & 88.79 & 87.29 & 93.94 & 92.15 & 94.34 & \textbf{95.32} \\ 
			~ & AA(\%) & 59.83 & 63.14 & 75.84 & 63.81 & 74.09 & 78.50 & 81.48 & \textbf{86.29} \\ 
			~ & Kappa(\%) & 78.80 & 77.94 & 85.18 & 82.75 & 92.00 & 89.56 & 92.51 & \textbf{93.82} \\
			\bottomrule[1.2pt]
	\end{tabular}}
\end{table*}

\begin{figure*}[htbp]
	\centering
	\includegraphics[scale=0.58]{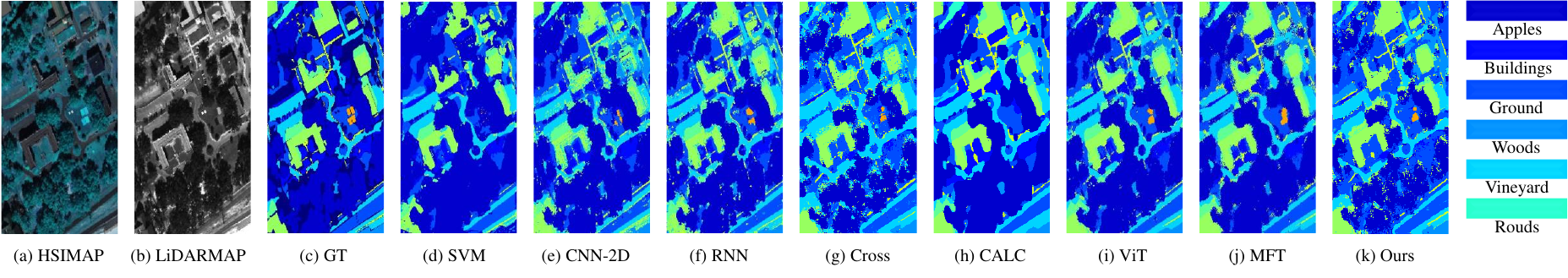}
	\centering
	\vspace{-0.2in}
	\caption{Visualization of false-color HSI and LiDAR images using different comparison methods based on the MUFFL dataset.}
	\vspace{-0.1in}
	\label{vis-m}
\end{figure*}

\subsection{Result Visualization}

To demonstrate the superiority of FedDiff in remote sensing multi-modal land classification tasks more intuitively, we adopted a visualization technique that assigns a unique color to each category. Firstly, we visualized the classification results of individual categories, specifically include categories 12 (Park lot 1), 14 (Tennis court), and 15 (Running track) on the Houston2013 dataset, as shown in Figure \ref{hotmap}. This visualization demonstrates FedDiff's clear classification performance on individual categories. Additionally, we present the classification visualization results on the three datasets in Figure \ref{vis-h}, Figure \ref{vis-t}, and Figure \ref{vis-m}. Traditional classifiers yield varying classification outcomes, and their multi-modal fusion methods solely rely on spectral representations of HSI data, lacking spatial information. As a result, these methods manifest prominent salt-and-pepper noise in the classification maps. In contrast, Transformer (ViT, MFT), with there network architecture emphasizing correlations between adjacent spectra and channels, as well as the ability to transmit position information across layers, produces visually superior classification maps. FedDiff integrates multi-modal heterogeneous data from different clients through an efficient federation framework. By combining frequency domain information to reconstruct high-dimensional features, it effectively reducing the granularity of textures in classification maps, resulting in more diverse and finer express. Overall, FedDiff efficiently performs feature-level fusion of heterogeneous data from various devices, while maintaining low communication cost, thus demonstrating superior results in land use and scene classification applications.

\section{Conclusion}
With the increasing attention on distributed remote sensing data platforms based on deep learning, multi-clients and multi-modal data fusion has become an indispensable direction.  Establishing a unified lightweight diffusion model without losing the manifold data of heterogeneous data is crucial to solve the federated feature fusion problem. In this regard, we propose a multi-modal collaborative diffusion federated learning framework called FedDiff. This framework effectively addresses the challenges of multi-clients and multi-modal fusion in remote sensing object classification tasks, without the need to expose privacy risks due to the transmission of redundant raw data or hinder communications. Extensive experiments are conducted on existing distributed hardware systems, demonstrating exceptional performance on online multi-modal data. We believe that the federated multi-modal learning framework will provide valuable insights for distributed remote sensing tasks.

\newpage
\bibliographystyle{IEEEtran}
\bibliography{reference}
\newpage

\vfill

\end{document}